# Variational Lossy Autoencoder

**Xi Chen**[†‡]**, Diederik P. Kingma**[‡]**, Tim Salimans**[‡]**, Yan Duan**[†‡]**, Prafulla Dhariwal**[‡]**,
John Schulman**[†‡]**, Ilya Sutskever**[‡]**, Pieter Abbeel**[†‡]
[†] UC Berkeley, Department of Electrical Engineering and Computer Science
[‡] OpenAI
`{peter,dpkingma,tim,rocky,prafulla,joschu,ilyasu,pieter}@openai.com`

## Abstract

Representation learning seeks to expose certain aspects of observed data in a learned representation that's amenable to downstream tasks like classification. For instance, a good representation for 2D images might be one that describes only global structure and discards information about detailed texture. In this paper, we present a simple but principled method to learn such global representations by combining Variational Autoencoder (VAE) with neural autoregressive models such as RNN, MADE and PixelRNN/CNN. Our proposed VAE model allows us to have control over what the global latent code can learn and by designing the architecture accordingly, we can force the global latent code to discard irrelevant information such as texture in 2D images, and hence the VAE only "autoencodes" data in a lossy fashion. In addition, by leveraging autoregressive models as both prior distribution $p(z)$ and decoding distribution $p(x|z)$, we can greatly improve generative modeling performance of VAEs, achieving new state-of-the-art results on MNIST, OMNIGLOT and Caltech-101 Silhouettes density estimation tasks as well as competitive results on CIFAR10.

## 1 Introduction

A key goal of representation learning is to identify and disentangle the underlying causal factors of the data, so that it becomes easier to understand the data, to classify it, or to perform other tasks (Bengio et al., 2013). For image data this often means that we are interested in uncovering the "global structure" that captures the content of an image (for example, the identity of objects present in the image) and its "style", but that we are typically less interested in the local and high frequency sources of variation such as the specific textures or white noise patterns.

A popular approach for learning representations is to fit a probabilistic latent variable model, an approach also known as *analysis-by-synthesis* (Yuille & Kersten, 2006; Nair et al., 2008). By learning a generative model of the data with the appropriate hierarchical structure of latent variables, it is hoped that the model will somehow uncover and untangle those causal sources of variations that we happen to be interested in. However, without further assumptions, representation learning via generative modeling is ill-posed: there are many different possible generative models with different (or no) kinds of latent variables that all encode the same probability density function on our observed data. Thus, the results we empirically get using this approach are highly dependent on the specific architectural and modeling choices that are made. Moreover, the objective that we optimize is often completely disconnected from the goal of learning a good representation: An autoregressive model of the data may achieve the same log-likelihood as a variational autoencoder (VAE) (Kingma & Welling, 2013), but the structure learned by the two models is completely different: the latter typically has a clear hierarchy of latent variables, while the autoregressive model has no stochastic latent variables at all (although it is conceivable that the deterministic hidden units of the autoregressive models will have meaningful and useful representations). For this reason, autoregressive models have thus far not been popular for the purpose of learning representations, even though they are extremely powerful as generative models (see e.g. van den Oord et al., 2016a).

A natural question becomes: is it possible to have a model that is a powerful density estimator and at the same time has the right hierarchical structure for representation learning? A potential solution would be to use a hybrid model that has both the latent variable structure of a VAE, as well as the powerful recurrence of an autoregressive model. However, earlier attempts at combining these two kinds of models have run into the problem that the autoregressive part of the model ends up explaining all structure in the data, while the latent variables are not used (Fabius & van Amersfoort, 2014; Chung et al., 2015; Bowman et al., 2015; Serban et al., 2016; Fraccaro et al., 2016; Xu & Sun, 2016). Bowman et al. (2015) noted that weakening the autoregressive part of the model by, for example, dropout can encourage the latent variables to be used. We analyze why weakening is necessary, and we propose a principled solution that takes advantage of this property to control what kind of information goes into latent variables. The model we propose performs well as a density estimator, as evidenced by state-of-the-art log-likelihood results on MNIST, OMNIGLOT and Caltech-101, and also has a structure that is uniquely suited for learning interesting global representations of data.

## 2 VAEs do not Autoencode in General

A VAE is frequently interpreted as a regularized autoencoder (Kingma & Welling, 2013; Zhang et al., 2016), but the conditions under which it is guaranteed to autoencode (reconstruction being close to original datapoint) are not discussed. In this section, we discuss the often-neglected fact that VAEs do not always autoencode and give explicit reasons why previous attempts to apply VAE in sequence modeling found that the latent code is generally not used unless the decoder is weakened (Bowman et al., 2015; Serban et al., 2016; Fraccaro et al., 2016). The understanding of when VAE does autoencode will be an essential building piece for VLAE.

### 2.1 Technical Background

Let $\mathbf{x}$ be observed variables, $\mathbf{z}$ latent variables and let $p(\mathbf{x}, \mathbf{z})$ be the parametric model of their joint distribution, called the *generative model* defined over the variables. Given a dataset $\mathbf{X} = \{\mathbf{x}^1, ..., \mathbf{x}^N\}$ we wish to perform maximum likelihood learning of its parameters:

$$\log p(\mathbf{X}) = \sum_{i=1}^{N} \log p(\mathbf{x}^{(i)}), \tag{1}$$

but in general this marginal likelihood is intractable to compute or differentiate directly for flexible generative models that have high-dimensional latent variables and flexible priors and likelihoods. A solution is to introduce $q(\mathbf{z}|\mathbf{x})$, a parametric *inference model* defined over the latent variables, and optimize the *variational lower bound* on the marginal log-likelihood of each observation $\mathbf{x}$:

$$\log p(\mathbf{x}) \geq \mathbb{E}_{q(\mathbf{z}|\mathbf{x})} \left[ \log p(\mathbf{x}, \mathbf{z}) - \log q(\mathbf{z}|\mathbf{x}) \right] = \mathcal{L}(\mathbf{x}; \theta) \tag{2}$$

where $\theta$ indicates the parameters of $p$ and $q$ models.

There are various ways to optimize the lower bound $\mathcal{L}(\mathbf{x}; \theta)$; for continuous $\mathbf{z}$ it can be done efficiently through a re-parameterization of $q(\mathbf{z}|\mathbf{x})$ (Kingma & Welling, 2013; Rezende et al., 2014).

This way of optimizing the variational lower bound with a parametric inference network and re-parameterization of continuous latent variables is usually called VAE. The "autoencoding" terminology comes from the fact that the lower bound $\mathcal{L}(\mathbf{x}; \theta)$ can be re-arranged:

$$\mathcal{L}(\mathbf{x}; \theta) = \mathbb{E}_{q(\mathbf{z}|\mathbf{x})} \left[ \log p(\mathbf{x}, \mathbf{z}) - \log q(\mathbf{z}|\mathbf{x}) \right] \tag{3}$$

$$= \mathbb{E}_{q(\mathbf{z}|\mathbf{x})} \left[ \log p(\mathbf{x}|\mathbf{z}) \right] - D_{KL}(q(\mathbf{z}|\mathbf{x})||p(\mathbf{z})) \tag{4}$$

where the first term can be seen as the expectation of negative reconstruction error and the KL divergence term can be seen as a regularizer, which as a whole could be seen as a regularized autoencoder loss with $q(\mathbf{z}|\mathbf{x})$ being the encoder and $p(\mathbf{x}|\mathbf{z})$ being the decoder. In the context of 2D images modeling, the decoding distribution $p(\mathbf{x}|\mathbf{z})$ is usually chosen to be a simple factorized distribution, i.e. $p(\mathbf{x}|\mathbf{z}) = \prod_i p(\mathbf{x}_i|\mathbf{z})$, and this setup often yields a sharp decoding distribution $p(\mathbf{x}|\mathbf{z})$ that tends to reconstruct original datapoint $\mathbf{x}$ exactly.

### 2.2 Bits-back Coding and Information preference

It's straightforward to see that having a more powerful $p(\mathbf{x}|\mathbf{z})$ will make VAE's marginal generative distribution $p(\mathbf{x}) = \int_{\mathbf{z}} p(\mathbf{z})p(\mathbf{x}|\mathbf{z})d\mathbf{z}$ more expressive. This idea has been explored extensively in previous work applying VAE to sequence modeling (Fabius & van Amersfoort, 2014; Chung et al., 2015; Bowman et al., 2015; Serban et al., 2016; Fraccaro et al., 2016; Xu & Sun, 2016), where the decoding distribution is a powerful RNN with autoregressive dependency, i.e., $p(\mathbf{x}|\mathbf{z}) = \prod_i p(\mathbf{x}_i|\mathbf{z}, \mathbf{x}_{<i})$. Since RNNs are universal function approximators and any joint distribution over $\mathbf{x}$ admits an autoregressive factorization, the RNN autoregressive decoding distribution can in theory represent any probability distribution even without dependence on $\mathbf{z}$.

However, previous attempts have found it hard to benefit from VAE when using an expressive decoding distribution $p(\mathbf{x}|\mathbf{z})$. Indeed it's documented in detail by Bowman et al. (2015) that in most cases when an RNN autoregressive decoding distribution is used, the latent code $\mathbf{z}$ is completely ignored and the model regresses to be a standard unconditional RNN autoregressive distribution that doesn't depend on the latent code. This phenomenon is commonly attributed to "optimization challenges" of VAE in the literature (Bowman et al., 2015; Serban et al., 2016; Kaae Sønderby et al., 2016) because early in the training the approximate posterior $q(\mathbf{z}|\mathbf{x})$ carries little information about datapoint $\mathbf{x}$ and hence it's easy for the model to just set the approximate posterior to be the prior to avoid paying any regularization cost $D_{KL}(q(\mathbf{z}|\mathbf{x})||p(\mathbf{z}))$.

Here we present a simple but often-neglected observation that this phenomenon arises not just due to optimization challenges and instead even if we can solve the optimization problems exactly, the latent code should still be ignored at optimum for most practical instances of VAE that have intractable true posterior distributions and sufficiently powerful decoders. It is easiest to understand this observation from a Bits-Back Coding perspective of VAE.

It is well-known that Bits-Back Coding is an information-theoretic view of Variational Inference (Hinton & Van Camp, 1993; Honkela & Valpola, 2004) and specific links have been established between Bits-Back Coding and the Helmholtz Machine/VAE (Hinton & Zemel, 1994; Gregor et al., 2013). Here we briefly relate VAE to Bits-Back Coding for self-containedness.

First recall that the goal of designing an efficient coding protocol is to minimize the expected code length of communicating $\mathbf{x}$. To explain Bits-Back Coding, let's first consider a more naive coding scheme. VAE can be seen as a way to encode data in a two-part code: $p(\mathbf{z})$ and $p(\mathbf{x}|\mathbf{z})$, where $\mathbf{z}$ can be seen as the essence/structure of a datum and is encoded first and then the modeling error (deviation from $\mathbf{z}$'s structure) is encoded next. The expected code length under this naive coding scheme for a given data distribution is hence:

$$\mathcal{C}_{\text{naive}}(\mathbf{x}) = \mathbb{E}_{\mathbf{x}\sim\text{data},\mathbf{z}\sim q(\mathbf{z}|\mathbf{x})}\left[-\log p(\mathbf{z}) - \log p(\mathbf{x}|\mathbf{z})\right] \tag{5}$$

This coding scheme is, however, inefficient. Bits-Back Coding improves on it by noticing that the encoder distribution $q(\mathbf{z}|\mathbf{x})$ can be used to transmit additional information, up to $H(q(\mathbf{z}|\mathbf{x}))$ expected nats, as long as the receiver also has access to $q(\mathbf{z}|\mathbf{x})$. The decoding scheme works as follows: a receiver first decodes $\mathbf{z}$ from $p(\mathbf{z})$, then decodes $\mathbf{x}$ from $p(\mathbf{x}|\mathbf{z})$ and, by running the same approximate posterior that the sender is using, decodes a secondary message from $q(\mathbf{z}|\mathbf{x})$. Hence, to properly measure the code length of VAE's two-part code, we need to subtract the extra information from $q(\mathbf{z}|\mathbf{x})$. Using Bit-Back Coding, the expected code length equates to the negative variational lower bound or the so-called Helmholtz variational free energy, which means minimizing code length is equivalent to maximizing the variational lower bound:

$$\mathcal{C}_{\text{BitsBack}}(\mathbf{x}) = \mathbb{E}_{\mathbf{x}\sim\text{data},\mathbf{z}\sim q(\mathbf{z}|\mathbf{x})}\left[\log q(\mathbf{z}|\mathbf{x}) - \log p(\mathbf{z}) - \log p(\mathbf{x}|\mathbf{z})\right] \tag{6}$$

$$= \mathbb{E}_{\mathbf{x}\sim\text{data}}\left[-\mathcal{L}(\mathbf{x})\right] \tag{7}$$

Casting the problem of optimizing VAE into designing an efficient coding scheme easily allows us to reason *when* the latent code $\mathbf{z}$ will be used: *the latent code $\mathbf{z}$ will be used when the two-part code is an efficient code.* Recalling that the lower-bound of expected code length for data is given by the Shannon entropy of data generation distribution: $\mathcal{H}(\text{data}) = \mathbb{E}_{x\sim\text{data}}\left[-\log p_{\text{data}}(x)\right]$, we can analyze VAE's coding efficiency:

$$\mathcal{C}_{\text{BitsBack}}(\mathbf{x}) = \mathbb{E}_{\mathbf{x}\sim\text{data},\mathbf{z}\sim q(\mathbf{z}|\mathbf{x})}\left[\log q(\mathbf{z}|\mathbf{x}) - \log p(\mathbf{z}) - \log p(\mathbf{x}|\mathbf{z})\right] \tag{8}$$

$$= \mathbb{E}_{\mathbf{x}\sim\text{data}}\left[-\log p(\mathbf{x}) + D_{KL}(q(\mathbf{z}|\mathbf{x})||p(\mathbf{z}|\mathbf{x}))\right] \tag{9}$$

$$\geq \mathbb{E}_{\mathbf{x}\sim\text{data}}\left[-\log p_{\text{data}}(\mathbf{x}) + D_{KL}(q(\mathbf{z}|\mathbf{x})||p(\mathbf{z}|\mathbf{x}))\right] \tag{10}$$

$$= \mathcal{H}(\text{data}) + \mathbb{E}_{\mathbf{x}\sim\text{data}}\left[D_{KL}(q(\mathbf{z}|\mathbf{x})||p(\mathbf{z}|\mathbf{x}))\right] \tag{11}$$

Since Kullback Leibler divergence is always non-negative, we know that using the two-part code derived from VAE suffers at least an extra code length of $D_{KL}(q(\mathbf{z}|\mathbf{x})||p(\mathbf{z}|\mathbf{x}))$ nats for using a posterior that's not precise. Many previous works in Variational Inference have designed flexible approximate posteriors to better approximate true posterior (Salimans et al., 2014; Rezende & Mohamed, 2015; Tran et al., 2015; Kingma et al., 2016). Improved posterior approximations have shown to be effective in improving variational inference but none of the existing methods are able to completely close the gap between approximate posterior and true posterior. This leads us to believe that for most practical models, at least in the near future, the extra coding cost $D_{KL}(q(\mathbf{z}|\mathbf{x})||p(\mathbf{z}|\mathbf{x}))$ will exist and will not be negligible.

Once we understand the inefficiency of the Bits-Back Coding mechanism, it's simple to realize why sometimes the latent code $z$ is not used: if the $p(\mathbf{x}|\mathbf{z})$ could model $p_{\text{data}}(\mathbf{x})$ without using information from $\mathbf{z}$, then it will not use $\mathbf{z}$, in which case the true posterior $p(\mathbf{z}|\mathbf{x})$ is simply the prior $p(\mathbf{z})$ and it's usually easy to set $q(\mathbf{z}|\mathbf{x})$ to be $p(\mathbf{z})$ to avoid incurring an extra cost $D_{KL}(q(\mathbf{z}|\mathbf{x})||p(\mathbf{z}|\mathbf{x}))$. And it's exactly the case when a powerful decoding distribution is used like an RNN autoregressive distribution, which given enough capacity is able to model arbitrarily complex distributions. Hence there exists a preference of information when a VAE is optimized: *information that can be modeled locally by decoding distribution $p(\mathbf{x}|\mathbf{z})$ without access to $\mathbf{z}$ will be encoded locally and only the remainder will be encoded in $\mathbf{z}$.*

We note that one common way to encourage putting information into the code is to use a factorized decoder $p(\mathbf{x}|\mathbf{z}) = \prod_i p(\mathbf{x}_i|\mathbf{z})$ but so long as there is one dimension $\mathbf{x}_j$ that's independent of all other dimensions for true data distribution, $p_{\text{data}}(\mathbf{x}) = p_{\text{data}}(\mathbf{x}_j)p_{\text{data}}(\mathbf{x}_{\neq j})$, then the latent code doesn't contain all the information about $\mathbf{x}$ since at least $\mathbf{x}_j$ will be modeled locally by factorized $p(\mathbf{x}|\mathbf{z})$. This kind of independence structure rarely exists in images so common VAEs that have factorized decoder autoencode almost exactly. Other techniques to encourage the usage of the latent code include annealing the relative weight of of $D_{KL}(q(\mathbf{z}|\mathbf{x})||p(\mathbf{z}))$ in the variational lower bound (Bowman et al., 2015; Kaae Sønderby et al., 2016) or the use of *free bits* (Kingma et al., 2016), which can serve the dual purpose of smoothing the optimization landscape and canceling out part of the Bits-Back Code inefficiency $D_{KL}(q(\mathbf{z}|\mathbf{x})||p(\mathbf{z}|\mathbf{x}))$.

## 3 Variational Lossy Autoencoder

The discussion in Section 2.2 suggests that autoregressive models cannot be combined with VAE since information will be preferred to be modeled by autoregressive models. Nevertheless, in this section, we present two complementary classes of improvements to VAE that utilize autoregressive models fruitfully to explicitly control representation learning and improve density estimation.

### 3.1 Lossy code via explicit information placement

Even though the information preference property of VAE might suggest that one should always use the full autoregressive models to achieve a better code length/log-likelihood, especially when slow data generation is not a concern, we argue that this information preference property can be exploited to turn the VAE into a powerful representation learning method that gives us fine-grained control over the kind of information that gets included in the learned representation.

When we try to learn a lossy compression/representation of data, we can simply construct a decoding distribution that's capable of modeling the part of information that we don't want the lossy representation to capture, but, critically, that's incapable of modelling the information that we do want the lossy representation to capture.

For instance, if we are interested in learning a global representation for 2D images that doesn't encode information about detailed texture, we can construct a specific factorization of the autoregressive distribution such that it has a small local receptive field as decoding distribution, e.g., $p_{\text{local}}(\mathbf{x}|\mathbf{z}) = \prod_i p(\mathbf{x}_i|\mathbf{z}, \mathbf{x}_{\text{WindowAround}(i)})$. Notice that, as long as $\mathbf{x}_{\text{WindowAround}(i)}$ is smaller than $\mathbf{x}_{<i}$, $p_{\text{local}}(\mathbf{x}|\mathbf{z})$ won't be able to represent arbitrarily complex distribution over $\mathbf{x}$ without dependence on $\mathbf{z}$ since the receptive field is limited such that not all distributions over $\mathbf{x}$ admit such factorizations. In particular, the receptive field window can be a small rectangle adjacent to a pixel $\mathbf{x}_i$ and in this case long-range dependency will be encoded in the latent code $\mathbf{z}$. On the other hand, if the true data distribution admits such factorization for a given datum $\mathbf{x}$ and dimension $i$, i.e.

$p_{\text{data}}(\mathbf{x}_i | \mathbf{x}_{\text{WindowAround}(i)}) = p_{\text{data}}(\mathbf{x}_i | \mathbf{x}_{<i})$, then the information preference property discussed in Section 2.2 will apply here, which means that all the information will be encoded in local autoregressive distribution for $\mathbf{x}_i$. Local statistics of 2D images like texture will likely be modeled completely by a small local window, whereas global structural information of an images like shapes of objects is long-range dependency that can only be communicated through latent code $\mathbf{z}$. Therefore we have given an example VAE that will produce a lossy compression of 2D images carrying exclusively global information that can't be modeled locally.

Notice that a global representation is only one of many possible lossy representations that we can construct using this information preference property. For instance, the conditional of an autoregressive distribution might depend on a heavily down-sampled receptive field so that it can only model long-range pattern whereas local high-frequency statistics need to be encoded into the latent code. Hence we have demonstrated that we can achieve explicit placement of information by constraining the receptive field/factorization of an autoregressive distribution that's used as decoding distribution.

We want to additionally emphasize the information preference property is an asymptotic view in a sense that it only holds when the variational lowerbound can be optimized well. Thus, we are not proposing an alternative to techniques like *free bits* Kingma et al. (2016) or KL annealing, and indeed they are still useful methods to smooth the optimization problem and used in this paper's experiments.

## 3.2 Learned Prior with Autoregressive Flow

Inefficiency in Bits-Back Coding, i.e., the mismatch between approximate posterior and true posterior, can be exploited to construct a lossy code but it's still important to minimize such inefficiency to improve overall modeling performance/coding efficiency. We propose to parametrize the prior distribution $p(\mathbf{z}; \theta)$ with an autoregressive model and show that a type of autoregressive latent code can in theory reduce inefficiency in Bits-Back Coding.

It is well-known that limited approximate posteriors impede learning and therefore various expressive posterior approximations have been proposed to improve VAE's density estimation performance (Turner et al., 2008; Mnih & Gregor, 2014; Salimans et al., 2014; Rezende & Mohamed, 2015; Kingma et al., 2016). One such class of approximate posteriors that has been shown to attain good empirical performance is based on the idea of Normalizing Flow, which is to apply an invertible mapping to a simple random variable, for example a factorized Gaussian as commonly used for $q(\mathbf{z}|\mathbf{x})$, in order to obtain a complicated random variable. For an invertible transformation between a simple distribution $\mathbf{y}$ and a more flexible $\mathbf{z}$, we know from the change-of-variable technique that $\log q(\mathbf{z}|\mathbf{x}) = \log q(\mathbf{y}|\mathbf{x}) - \log \det \frac{d\mathbf{z}}{d\mathbf{y}}$ and using $q(\mathbf{z}|\mathbf{x})$ as approximate posterior will decrease the coding efficiency gap $D_{KL}(q(\mathbf{z}|\mathbf{x})||p(\mathbf{z}|\mathbf{x}))$ provided the transformation is sufficiently expressive.

Kingma et al. (2016) introduced Inverse Autoregressive Flow, which is a powerful class of such invertible mappings that have simple determinant: $z_i = \frac{y_i - \mu_i(y_{1:i-1})}{\sigma_i(y_{1:i-1})}$, where $\mu_i(.) \in \mathcal{R}, \sigma_i(.) \in \mathcal{R}^+$ are general functions that can be parametrized by expressive neural networks, such as MADE and PixelCNN variants (Germain et al., 2015; van den Oord et al., 2016a). Inverse autoregressive flow is the inverse/whitening of autoregressive flow: $y_i = z_i \sigma_i(y_{1:i-1}) + \mu_i(y_{1:i-1})$. We refer interested readers to (Rezende & Mohamed, 2015; Kingma et al., 2016) for in-depth discussions on related topics.

In this paper, we propose to parametrize our learnable prior as an autoregressive flow from some simple noise source like spherical Gaussian. Next, we show that using latent code transformed by autoregressive flow (AF) is equivalent to using inverse autoregressive flow (IAF) approximate posterior, which explains why it can similarly improve Bits-Back Coding efficiency. Moreover, compared with an IAF posterior, an AF prior has a more expressive generative model that essentially "comes for free".

For an autoregressive flow $f$, some continuous noise source $\epsilon$ is transformed into latent code $\mathbf{z}$: $\mathbf{z} = f(\epsilon)$. Assuming the density function for noise source is $u(\epsilon)$, we similarly know that $\log p(\mathbf{z}) = \log u(\epsilon) + \log \det \frac{d\epsilon}{d\mathbf{z}}$.

Simply re-arranging the variational lowerbound for using AF prior reveals that having an AF latent code $\mathbf{z}$ is equivalent to using an IAF posterior for $\epsilon$ that we can interpret as the new latent code:

$$\mathcal{L}(\mathbf{x}; \theta) = \mathbb{E}_{\mathbf{z} \sim q(\mathbf{z}|\mathbf{x})} \left[ \log p(\mathbf{x}|\mathbf{z}) + \log p(\mathbf{z}) - \log q(\mathbf{z}|\mathbf{x}) \right] \tag{12}$$

$$= \mathbb{E}_{\mathbf{z} \sim q(\mathbf{z}|\mathbf{x}), \epsilon = f^{-1}(\mathbf{z})} \left[ \log p(\mathbf{x}|f(\epsilon)) + \log u(\epsilon) + \log \det \frac{d\epsilon}{d\mathbf{z}} - \log q(\mathbf{z}|\mathbf{x}) \right] \tag{13}$$

$$= \mathbb{E}_{\mathbf{z} \sim q(\mathbf{z}|\mathbf{x}), \epsilon = f^{-1}(\mathbf{z})} \left[ \log p(\mathbf{x}|f(\epsilon)) + \log u(\epsilon) - \underbrace{\left( \log q(\mathbf{z}|\mathbf{x}) - \log \det \frac{d\epsilon}{d\mathbf{z}} \right)}_{\text{IAF Posterior}} \right] \tag{14}$$

AF prior is the same as IAF posterior along the encoder path, $f^{-1}(q(\mathbf{z}|\mathbf{x}))$, but differs along the decoder/generator path: IAF posterior has a shorter decoder path $p(\mathbf{x}|\mathbf{z})$ whereas AF prior has a deeper decoder path $p(\mathbf{x}|f(\epsilon))$. The crucial observation is that AF prior and IAF posterior have the same computation cost under the expectation of $\mathbf{z} \sim q(\mathbf{z}|\mathbf{x})$, so using AF prior makes the model more expressive at no training time cost.

## 4 EXPERIMENTS

In this paper, we evaluate VLAE on 2D images and leave extensions to other forms of data to future work. For the rest of the section, we define a VLAE model as a VAE that uses AF prior and autoregressive decoder. We choose to implement conditional distribution $p(\mathbf{x}|\mathbf{z})$ with a small-receptive-field PixelCNN (van den Oord et al., 2016a), which has been proved to be a scalable autoregressive model.

For evaluation, we use binary image datasets that are commonly used for density estimation tasks: MNIST (LeCun et al., 1998) (both statically binarized [1] and dynamically binarized version (Burda et al., 2015a)), OMNIGLOT (Lake et al., 2013; Burda et al., 2015a) and Caltech-101 Silhouettes (Marlin et al., 2010). All datasets uniformly consist of 28x28 binary images, which allow us to use a unified architecture. VAE networks used in binary image datasets are simple variants of ResNet VAEs described in (Salimans et al., 2014; Kingma et al., 2016). For the decoder, we use a variant of PixelCNN that has 6 layers of masked convolution with filter size 3, which means the window of dependency, $\mathbf{x}_{\text{WindowAround}(i)}$, is limited to a small local patch. During training, "free bits" (Kingma et al., 2016) is used improve optimization stability. Experimental setup and hyperparameters are detailed in the appendix. Reported marginal NLL is estimated using Importance Sampling with 4096 samples.

We designed experiments to answer the following questions:

- Can VLAE learn lossy codes that encode global statistics?
- Does using AF priors improves upon using IAF posteriors as predicted by theory?
- Does using autoregressive decoding distributions improve density estimation performance?

### 4.1 LOSSY COMPRESSION

First we are interested in whether VLAE can learn a lossy representation/compression of data by using the PixelCNN decoder to model local statistics. We trained VLAE model on Statically Binarized MNIST and the converged model has $\mathbb{E}[D_{KL}(q(\mathbf{z}|\mathbf{x})||p(\mathbf{z}))] = 13.3$ nats $= 19.2$ bits, which is the number of bits it uses on average to encode/compress one MNIST image. By comparison, an identical VAE model with factorized decoding distribution will uses on average 37.3 bits in latent code, and this thus indicates that VLAE can learn a lossier compression than a VAE with regular factorized conditional distribution.

The next question is whether VLAE's lossy compression encodes global statistics and discards local statistics. In Fig 1a, we visualize original images $\mathbf{x}_{\text{data}}$ and one random "decompression" $\mathbf{x}_{\text{decompressed}}$ from VLAE: $\mathbf{z} \sim q(\mathbf{z}|\mathbf{x}_{\text{data}}), \mathbf{x}_{\text{decompressed}} \sim p(\mathbf{x}|\mathbf{z})$. We observe that none of the

---

[1] We use the version provided by Hugo Larochelle.

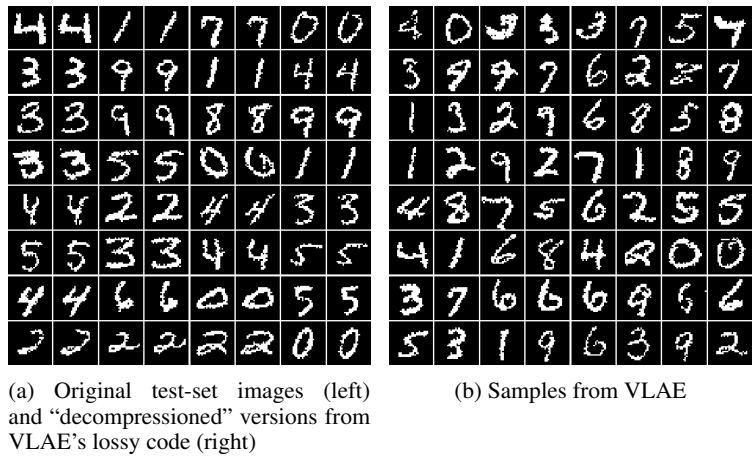

(a) Original test-set images (left) and "decompressioned" versions from VLAE's lossy code (right)

(b) Samples from VLAE

Figure 1: Statically Binarized MNIST

decompressions is an exact reconstruction of the original image but instead the global *structure* of the image was encoded in the lossy code **z** and regenerated. Also worth noting is that local statistics are not preserved but a new set of likely local statistics are generated in the decompressed images: the binary masks are usually different and local styles like stroke width are sometimes slightly different.

However, we remark that the lossy code **z** doesn't always capture the kind of global information that we care about and it's dependent on the type of constraint we put on the decoder. For instance, in Fig 4b, we show decompressions for OMNIGLOT dataset, which has more meaningful variations in small patches than MNIST, and we can observe that semantics are not preserved in some cases. This highlights the need to *specify* the type of statistics we care about in a representation, which will be different across tasks and datasets, and design decoding distribution accordingly.

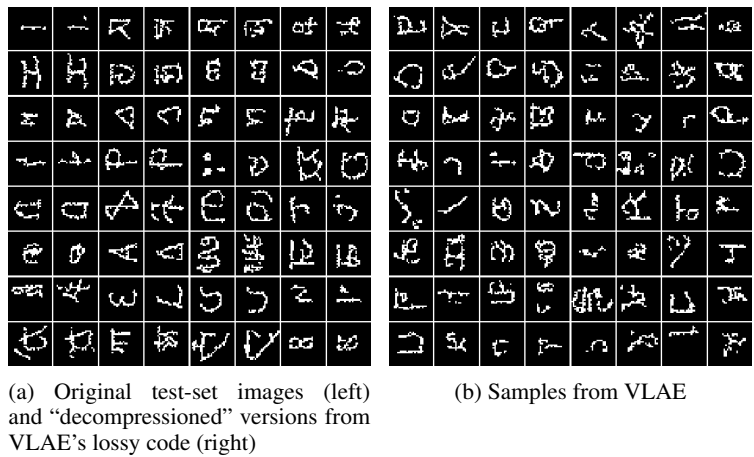

(a) Original test-set images (left) and "decompressioned" versions from VLAE's lossy code (right)

(b) Samples from VLAE

Figure 2: OMNIGLOT

## 4.2 Density Estimation

Next we investigate whether leveraging autoregressive models as latent distribution $p(\mathbf{z})$ and as decoding distribution $p(\mathbf{x}|\mathbf{z})$ would improve density estimation performance.

To verify whether AF prior is able to improve upon IAF posterior alone, it's desirable to test this model without using autoregressive decoder but instead using the conventional independent Bernoulli distribution for $p(\mathbf{x}|\mathbf{z})$. Hence we use the best performing model from Kingma et al.

Table 1: Statically Binarized MNIST

| Model | NLL Test |
|---|---|
| Normalizing flows (Rezende & Mohamed, 2015) | 85.10 |
| DRAW (Gregor et al., 2015) | < 80.97 |
| Discrete VAE (Rolfe, 2016) | 81.01 |
| PixelRNN (van den Oord et al., 2016a) | 79.20 |
| IAF VAE (Kingma et al., 2016) | 79.88 |
| AF VAE | 79.30 |
| VLAE | **79.03** |

(2016) on statically binarized MNIST and make the single modification of replacing the original IAF posterior with an equivalent AF prior, removing the context. As seen in Table 1, VAE with AF prior is outperforming VAE with an equivalent IAF posterior, indicating that the deeper generative model from AF prior is beneficial. A similar gain carries over when an autoregressive decoder is used: on statically binarized MNIST, using AF prior instead of IAF posterior reduces train NLL by 0.8 nat and test NLL by 0.6 nat.

Next we evaluate whether using autoregressive decoding distribution can improve performance and we show in Table 1 that a VLAE model, with AF prior and PixelCNN conditional, is able to outperform a VAE with just AF prior and achieves new state-of-the-art results on statically binarized MNIST.

In addition, we hypothesize that the separation of different types of information, the modeling global structure in latent code and local statistics in PixelCNN, likely has some form of good inductive biases for 2D images. In order to evaluate if VLAE is an expressive density estimator with good inductive biases, we will test a single VLAE model, *with the same network architecture*, on all binary datasets. We choose hyperparameters manually on statically binarized MNIST and use the same hyperparameters to evaluate on dynamically binarized MNIST, OMNIGLOT and Caltech-101 Silhouettes. We also note that better performance can be obtained if we individually tune hyperparameters for each dataset. As a concrete demonstration, we report the performance of a fine-tuned VLAE on OMNIGLOT dataset in Table 3.

Table 2: Dynamically binarized MNIST

| Model | NLL Test |
|---|---|
| Convolutional VAE + HVI (Salimans et al., 2014) | 81.94 |
| DLGM 2hl + IWAE (Burda et al., 2015a) | 82.90 |
| Discrete VAE (Rolfe, 2016) | 80.04 |
| LVAE (Kaae Sønderby et al., 2016) | 81.74 |
| DRAW + VGP (Tran et al., 2015) | < 79.88 |
| IAF VAE (Kingma et al., 2016) | 79.10 |
| Unconditional Decoder | 87.55 |
| VLAE | **78.53** |

Table 3: OMNIGLOT. [1] (Burda et al., 2015a), [2] (Burda et al., 2015b), [3] (Gregor et al., 2015), [4] (Gregor et al., 2016),

| Model | NLL Test |
|---|---|
| VAE [1] | 106.31 |
| IWAE [1] | 103.38 |
| RBM (500 hidden) [2] | 100.46 |
| DRAW [3] | < 96.50 |
| Conv DRAW [4] | < 91.00 |
| Unconditional Decoder | 95.02 |
| VLAE | 90.98 |
| VLAE (fine-tuned) | **89.83** |

Table 4: Caltech-101 Silhouettes. [1] (Bornschein & Bengio, 2014), [2] (Cho et al., 2011), [3] (Du et al., 2015), [4] (Rolfe, 2016), [5] (Goessling & Amit, 2015),

| Model | NLL Test |
|---|---|
| RWS SBN [1] | 113.3 |
| RBM [2] | 107.8 |
| NAIS NADE [3] | 100.0 |
| Discrete VAE [4] | 97.6 |
| SpARN [5] | 88.48 |
| Unconditional Decoder | 89.26 |
| VLAE | **77.36** |

As seen in Table 2,3,4, with the same set of hyperparameters tuned on statically binarized MNIST, VLAE is able to perform well on the rest of datasets, significantly exceeding previous state-of-the-art results on dynamically binarized MNIST and Caltech-101 Silhouettes and tying statistically with best previous result on OMNIGLOT. In order to isolate the effect of expressive PixelCNN as decoder, we also report performance of the same PixelCNN trained without VAE part under the name "Unconditional Decoder".

### 4.3 NATURAL IMAGES: CIFAR10

In addition to binary image datasets, we have applied VLAE to the CIFAR10 dataset of natural images. Density estimation of CIFAR10 images has been a challenging benchmark problem used by many recent generative models and hence is great task to position VLAE among existing methods.

We investigated using ResNet (He et al., 2016) and DenseNet (Huang et al., 2016) as building blocks for VAE networks and observed that DenseNet reduces overfitting. We also propose a new optimization technique that blends the advantages of KL annealing (Serban et al., 2016) and "free bits" (Kingma et al., 2016) to stabilize learning on this challenging dataset. Detailed experimental setup is described in Appendix.

VLAE is compared to other methods on CIFAR10 in Table 5. We show that VLAE models attain new state-of-the-art performance among other variationally trained latent-variable models. DenseNet VLAE model also outperforms most other tractable likelihood models including Gated PixelCNN and PixelRNN and has results only slightly worse than currently unarchived state-of-the-art PixelCNN++.

Table 5: CIFAR10. Likelihood for VLAE is approximated with 512 importance samples. [1] (van den Oord et al., 2016a), [2] (Dinh et al., 2014), [3] (van den Oord & Schrauwen, 2014), [4] (Dinh et al., 2016), [5] (van den Oord et al., 2016b), [6] (Salimans et al., 2017), [7] (Sohl-Dickstein et al., 2015), [8] (Gregor et al., 2016), [9] (Kingma et al., 2016)

| Method | bits/dim $\leq$ |
|---|---|
| *Results with tractable likelihood models*: | |
| Uniform distribution [1] | 8.00 |
| Multivariate Gaussian [1] | 4.70 |
| NICE [2] | 4.48 |
| Deep GMMs [3] | 4.00 |
| Real NVP [4] | 3.49 |
| PixelCNN [1] | 3.14 |
| Gated PixelCNN [5] | 3.03 |
| PixelRNN [1] | 3.00 |
| PixelCNN++ [6] | **2.92** |
| *Results with variationally trained latent-variable models*: | |
| Deep Diffusion [7] | 5.40 |
| Convolutional DRAW [8] | 3.58 |
| ResNet VAE with IAF [9] | 3.11 |
| ResNet VLAE | 3.04 |
| DenseNet VLAE | **2.95** |

We also investigate learning lossy codes on CIFAR10 images. To illustrate how does the receptive field size of PixelCNN decoder influence properties of learned latent codes, we show visualizations of similar VLAE models with receptive fields of different sizes. Specifically we say a receptive field, $\mathbf{x}_{\text{WindowAround}(i)}$, has size $A x B$ when a pixel $x_i$ can depend on the rectangle block of size $A x B$ immediately on top of $x_i$ as well as the $\left\lceil \frac{A-1}{2} \right\rceil$ pixels immediately to the left of $x_i$. We use this notation to refer to different types of PixelCNN decoders in Figure 3.

From (a)-(c) in Figure 3, we can see that larger receptive fields progressively make autoregressive decoders capture more structural information. In (a), a smaller receptive field tends to preserve rather detailed shape information in the lossy code whereas the latent code only retains rough shape in (c) with a larger receptive field.

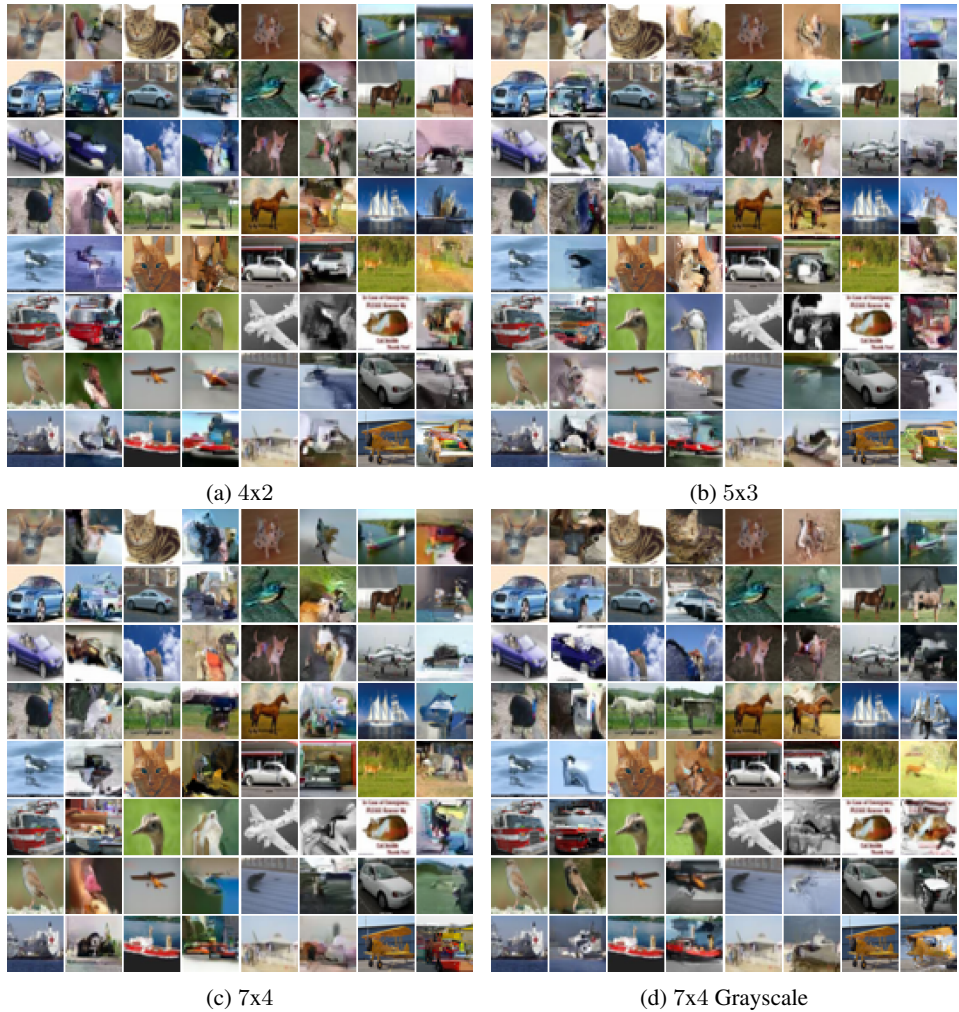

(a) 4x2

(b) 5x3

(c) 7x4

(d) 7x4 Grayscale

Figure 3: CIFAR10: Original test-set images (left) and "decompressioned" versions from VLAE's lossy code (right) with different types of receptive fields

It's interesting to also note that in (a)-(c), oftentimes color information is partially omitted from latent codes and one explanation can be that color is very predictable locally. However, color information can be important to preserve if our task is, for example, object classification. To demonstrate how we can encode color information in the lossy code, we can choose to make PixelCNN decoder depend only on images' grayscale versions. In other words, instead of choosing the decoder to be $p_{\text{local}}(\mathbf{x}|\mathbf{z}) = \prod_i p(\mathbf{x}_i|\mathbf{z}, \mathbf{x}_{\text{WindowAround}(i)})$, we use a decoder of the form $p_{\text{local}}(\mathbf{x}|\mathbf{z}) = \prod_i p(\mathbf{x}_i|\mathbf{z}, \text{Grayscale}(\mathbf{x}_{\text{WindowAround}(i)}))$. In (d) of Figure 3, we visualize lossy codes for a VLAE that has the same receptive field size as (c) but uses a "grayscale receptive field". We note that the lossy codes in (d) encode roughly the same structural information as those in (c) but generally generate objects that are more recognizable due to the preservation of color information. This serves as one example of how we can design the lossy latent code carefully to encode what's important and what's not.

## 5 RELATED WORK

We investigate a fusion between variational autoencoders with continuous latent variables (Kingma & Welling, 2013; Rezende et al., 2014) and neural autoregressive models. For autoregression, we specifically apply a novel type of architecture where autoregression is realised through a carefully constructed deep convolutional network, introduced in the PixelCNN model for images (van den Oord et al., 2016a,b). These family of convolutional autoregressive models was further explored, and extended, for audio in WaveNet (Oord et al., 2016), video in Video Pixel Networks (Kalchbrenner et al., 2016b) and language in ByteNet (Kalchbrenner et al., 2016a).

The combination of latent variables with expressive decoder was previously explored using recurrent networks mainly in the context of language modeling (Chung et al., 2015; Bowman et al., 2015; Serban et al., 2016; Fraccaro et al., 2016; Xu & Sun, 2016). Bowman et al. (2015) has also proposed to weaken an otherwise too expressive decoder by dropout to force some information into latent codes.

Concurrent with our work, PixelVAE (Gulrajani et al., 2016) also explored using conditional Pixel-CNN as a VAE's decoder and has obtained impressive density modeling results through the use of multiple levels of stochastic units.

Using autoregressive model on latent code was explored in the context of discrete latent variables in DARN (Gregor et al., 2013). Kingma et al. (2016), Kaae Sønderby et al. (2016), Gregor et al. (2016) and Salimans (2016) explored VAE architecture with an explicitly deep autoregressive prior for continuous latent variables, but the autoregressive data likelihood is intractable in those architectures and needs to inferred variationally. In contrast, we use multiple steps of autoregressive flows that has exact likelihood and analyze the effect of using expressive latent code.

Optimization challenges for using (all levels of) continuous latent code were discussed before and practical solutions were proposed (Bowman et al., 2015; Kaae Sønderby et al., 2016; Kingma et al., 2016). In this paper, we present a complementary perspective on when/how should the latent code be used by appealing to a Bits-Back interpretation of VAE.

Learning a lossy compressor with latent variable model has been investigated with Con-vDRAW (Gregor et al., 2016). It learns a hierarchy of latent variables and just using high-level latent variables will result in a lossy compression that performs similarly to JPEG. Our model similarly learns a lossy compressor but it uses an autoregressive model to explicitly control what kind of information should be lost in compression.

## 6 CONCLUSION

In this paper, we analyze the condition under which the latent code in VAE should be used, i.e. when does VAE autoencode, and use this observation to design a VAE model that's a lossy compressor of observed data. At modeling level, we propose two complementary improvements to VAE that are shown to have good empirical performance.

VLAE has the appealing properties of controllable representation learning and improved density estimation performance but these properties come at a cost: compared with VAE models that have simple prior and decoder, VLAE is slower at generation due to the sequential nature of autoregressive model.

Moving forward, we believe it's exciting to extend this principle of learning lossy codes to other forms of data, in particular those that have a temporal aspect like audio and video. Another promising direction is to design representations that contain only information for downstream tasks and utilize those representations to improve semi-supervised learning.

## APPENDIX

## A   DETAILED EXPERIMENT SETUP FOR BINARY IMAGES

For VAE's encoder and decoder, we use the same ResNet (He et al., 2015) VAE architecture as the one used in IAF MNIST experiment (Kingma et al., 2016). The only difference is that the decoder network now, instead of outputing a 28x28x1 spatial feature map to specify the mean of a factorized bernoulli distribution, outputs a 28x28x4 spatial feature map that's concatenated with the original binary image channel-wise, forming a 28x28x5 feature map that's then fed through a typical masked PixelCNN (van den Oord et al., 2016a). As such even though the PixelCNN conditions on the latent code, we don't call it a Conditional PixelCNN because it doesn't use the specific architecture that was proposed in van den Oord et al. (2016b). For the PixelCNN, it has 6 masked convolution layers with 12 3x3 filters organized in ResNet blocks, and it has 4 additional 1x1 convolution ResNet block between every other masked convolution layer to increase processing capacity since it employs fewer masked convolutions than usual. All the masked convolution layer have their weights tied to reduce overfitting on statically binarized MNIST, and untying the weights will increase performance for other datasets. Experiments are tuned on the validation set and then final experiment was run with train and validation set, with performance evaluated with test set. Exponential Linear Units (Clevert et al., 2015) are used as activation functions in both VAE network and PixelCNN network. Weight normalization is everywhere with data-dependent initialization (Salimans & Kingma, 2016).

A latent code of dimension $64$ was used. For AF prior, it's implemented with MADE (Germain et al., 2015) as detailed in Kingma et al. (2016). We used $4$ steps of autoregressive flow and each flow is implemented by a 3-layer MADE that has 640 hidden units and uses Relu (Nair & Hinton, 2010) as activation functions. Differing from the practice of Kingma et al. (2016), we use mean-only autoregressive flow, which we found to be more numerically stable.

In terms of training, Adamax (Kingma & Ba, 2014) was used with a learning rate of $0.002$. $0.01$ nats/data-dim free bits (Kingma et al., 2016) was found to be effective in dealing with the problem of all the latent code being ignored early in training. Polyak averaging (Polyak & Juditsky, 1992) was used to compute the final parameters, with $\alpha = 0.998$.

All experiments are implemented using TensorFlow (Abadi et al., 2016).

## B   ADDITIONAL EXPERIMENT SETUP FOR CIFAR10

Latent codes are represented by 16 feature maps of size $8x8$, and this choice of spatial stochastic units are inspired by ResNet IAF VAE (Kingma et al., 2016). Prior distribution is factorized Gaussian noise transformed by 6 autoregressive flows, each of which is implemented by a PixelCNN (van den Oord et al., 2016a) with 2 hidden layers and 128 feature maps. Between every other autoregressive flow, the ordering of stochastic units is reversed.

ResNet VLAE has the following structure for encoder: 2 ResNet blocks, Conv w/ stride=2, 2 ResNet blocks, Conv w/ stride=2, 3 ResNet blocks, 1x1 convolution and has a symmetric decoder. Channel size = 48 for 32x32 feature maps and 96 for other feature maps. DenseNet VLAE follows a similar structure: replacing 2 ResNet blocks with one DenseNet block of 3 steps and each step produces a certain number of feature maps such that at the end of a block, the concatenated feature maps is slightly more than the ResNet VLAE at the same stage.

Conditional PixelCNN++ (Salimans et al., 2017) is used as the decoder. Specifically the channel-autoregressive variant is used to ensure there is sufficient capacity even when the receptive field is small. Specifically, the decoder PixelCNN has 4 blocks of 64 feature maps where each block is conditioned on previous blocks with Gated ResNet connections and hence the PixelCNN decoders we use are shallow but very wide. For 4x2 receptive field experiment, we use 1 layer of vertical stack convolutions and 2 layers of horizontal stack convolutions; for 5x3 receptive field experiment, we use 2 layers of vertical stack convolutions and 2 layers of horizontal stack convolutions; For 5x3 receptive field experiment, we use 2 layers of vertical stack convolutions and 2 layers of horizontal stack convolutions; For 7x4 receptive field experiment, we use 3 layers of vertical stack convolutions and 3 layers of horizontal stack convolutions; for 7x4 Grayscale experiment, we transform RGB images into gray-scale images via this specific transformation: $(0.299 * R) + (0.587G) + (0.114B)$. Best density estimation result is obtained with 7x4 receptive field experiments.

## C  SOFT FREE BITS

"Free bits" was a technique proposed in (Kingma et al., 2016) where $K$ groups of stochastic units are encouraged to be used through the following surrogate objective:

$$\widetilde{\mathcal{L}}_\lambda = \mathbb{E}_{\mathbf{x} \sim \mathcal{M}} \left[ \mathbb{E}_{q(\mathbf{z}|\mathbf{x})} \left[ \log p(\mathbf{x}|\mathbf{z}) \right] \right] - \sum_{j=1}^{K} \text{maximum}(\lambda, \mathbb{E}_{\mathbf{x} \sim \mathcal{M}} \left[ D_{KL}(q(\mathbf{z}_j|\mathbf{x})||p(\mathbf{z}_j)) \right])$$

This technique is easy to use since it's usually easy to determine the minimum number of bits/nats, $\lambda$, stochastic units need to encode. Choosing $\lambda$ is hence easier than setting a fixed KL annealing schedule (Serban et al., 2016).

On the other hand, Kl annealing has the benefit of the surrogate objective will smoothly become the true objective, the variational lower bound where as "free bits" has a sharp transition at the boundary. Therefore, we propose to still use $\lambda$ as hyperparameter to specify at least $\lambda$ nats should be used but try to change the optimization objective as slowly as possible:

$$\mathcal{L}_{\text{SoftFreeBits}}(\mathbf{x}; \theta) = \mathbb{E}_{q(\mathbf{z}|\mathbf{x})} \left[ \log p(\mathbf{x}|\mathbf{z}) \right] - \gamma D_{KL}(q(\mathbf{z}|\mathbf{x})||p(\mathbf{z}))$$

where $0 < \gamma \leq 1$.

And we make the optimization smoother by changing $\gamma$ slowly online to make sure at least $\lambda$ nats are used: when Kl is too much higher than $\lambda$ (we experimented wide range of thresholds from 3% to 30%, all of which yield improved results, and we tend to use 5% us a threshold), $\gamma$ is increased, and when Kl lower than $\lambda$, $\gamma$ is decreased to encourage information flow.

We found it sufficient to increase/decrease at $10\%$ increment and didn't further tune this parameter.

## D  AUTOREGRESSIVE DECODER WITHOUT AUTOREGRESSIVE PRIOR

In this section, we investigate the scenario of just using an autoregressive decoder without using an autoregressive prior. We compare the exact same model in three configurations: 1) using small-receptive-field PixelCNN as an unconditional density estimator; 2) using small-receptive-field as a decoder in a VAE with Gaussian latent variables; 3) replacing Gaussian latent variables with autoregressive flow latent variables in 2).

Table 1: Ablation on Dynamically binarized MNIST

| Model | NLL Test | KL |
|---|---|---|
| Unconditional PixelCNN | 87.55 | 0 |
| PixelCNN Decoder + Gaussian Prior | 79.48 | 10.60 |
| PixelCNN Decoder + AF Prior | **78.94** | 11.73 |

In Table 1, we can observe that each step of modification improves density estimation performance. In addition, using an autoregressive latent code makes the latent code transmit more information as shown in the difference of $\mathbb{E}[D_{KL}(q(\mathbf{z}|\mathbf{x})||p(\mathbf{z}))]$.

## E  CIFAR10 GENERATED SAMPLES

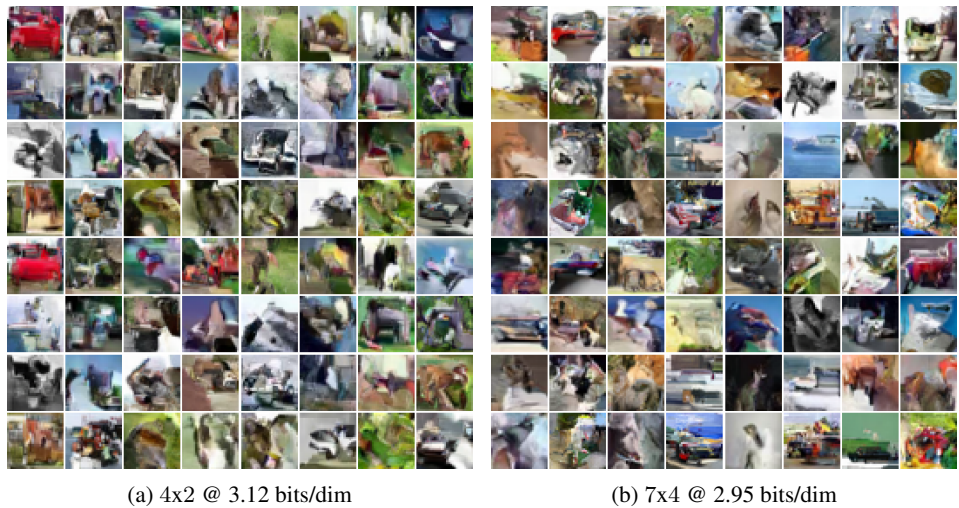

(a) 4x2 @ 3.12 bits/dim        (b) 7x4 @ 2.95 bits/dim

Figure 4: CIFAR10: Generated samples for different models